\documentclass[10pt,final]{IEEEtran}

\usepackage{amsmath}
\usepackage{amsfonts}
\usepackage{amssymb}
\usepackage{amsbsy}
\usepackage{epsfig}
\usepackage{graphicx}
\usepackage[latin1]{inputenc}
\usepackage{latexsym}
\usepackage{indentfirst}
\usepackage{fixltx2e}
\usepackage{wrapfig}
\usepackage{eepic}
\usepackage{lscape}
\usepackage[caption=false,font=footnotesize]{subfig}

\newcommand{\ww}{\mathbf{w}}

\newcommand{\WW}{\mathbf{W}}
\newcommand{\VV}{\mathbf{V}}
\newcommand{\UU}{\mathbf{U}}

\newcommand{\TT}{^\mathsf{T}}

\newcommand{\GG}{\boldsymbol{\mathit{\Gamma}}}

\newcommand{\ii}{\text{i}}

\begin{document}

\title{Source Separation and Clustering \\of Phase-Locked Subspaces: Derivations and Proofs}

\author{Miguel Almeida,
				Jan-Hendrik Schleimer,
				José Bioucas-Dias,
       	Ricardo Vigário
       	\thanks{Miguel Almeida is with the Institute of Telecommunications, Superior Technical Institute, Portugal. Email: malmeida@lx.it.pt}
       	\thanks{Jan-Hendrik Schleimer is with the Bernstein Center for Computational Neuroscience, Humboldt University, Berlin. Email: jan-hendrik.schleimer@bccn-berlin.de}
       	\thanks{José Bioucas-Dias is with the Institute of Telecommunications, Superior Technical Institute, Portugal. Email: bioucas@lx.it.pt}
       	\thanks{Ricardo Vigário is with the Adaptive Informatics Research Centre, Aalto University, Finland. Email: ricardo.vigario@hut.fi}}

\maketitle

\begin{abstract}
Due to space limitations, our submission ``Source Separation and Clustering of Phase-Locked Subspaces'', accepted for publication on the IEEE Transactions on Neural Networks in 2011, presented some results without proof. Those proofs are provided in this paper.
\end{abstract}



\appendices


\section{Gradient of $\left|\varrho\right|^2$ in RPA}
\label{AppRefGrad}

In this section we derive that the gradient of $|\varrho|^2$ is given by Eq. 6 of \cite{Almeida2011}, where $|\varrho|$ is defined as in Eq. 5 of \cite{Almeida2011}. Recall that $\Delta\phi(t) = \phi(t) - \psi(t)$, where $\phi(t)$ is the phase of the estimated source $y(t) = \ww^T\mathbf{x}(t)$ and $\psi(t)$ is the phase of the reference $u(t)$. Further, define $\varrho \equiv |\varrho|e^{\ii\Phi}$.

We begin by noting that $|\varrho|^2 = (|\varrho| \cos(\Phi))^2 + (|\varrho| \sin(\Phi))^2$, so that
\begin{align*}
\nabla|\varrho|^2 &= \nabla(|\varrho|\cos(\Phi))^2 + \nabla(|\varrho|\sin(\Phi))^2\\
&= 2|\varrho|\left[\cos(\Phi) \nabla(|\varrho|\cos(\Phi)) + \sin(\Phi)\nabla(|\varrho|\sin(\Phi))\right].
\end{align*}
Note that we have $\frac{1}{T}\sum_{t=1}^T \cos(\Delta\phi(t))) = |\varrho|\cos(\Phi)$ and $\frac{1}{T}\sum_{t=1}^T \sin(\Delta\phi(t))) = |\varrho|\sin(\Phi)$, so we get
\begin{align}
\nabla|\varrho|^2 &= 2|\varrho|\left\lbrace \cos(\Phi)\nabla\left[ \frac{1}{T}\sum_{t=1}^T \cos(\Delta\phi(t))) \right]\right. +  \nonumber \\
& \qquad + \left.\sin(\Phi)\nabla\left[ \frac{1}{T}\sum_{t=1}^T \sin(\Delta\phi(t))) \right] \right\rbrace \nonumber\\
&\hspace{-0.5cm}= \frac{2|\varrho|}{T} \sum_{t=1}^T \Big[\sin(\Phi) \cos(\Delta\phi(t)) - \cos(\Phi) \sin(\Delta\phi(t))\Big] \times \nonumber \\
& \qquad \times \nabla\Delta\phi(t) \nonumber\\
&\hspace{-0.5cm}= \frac{2|\varrho|}{T} \sum_{t=1}^T \sin[\Phi - \Delta\phi(t)] \nabla\Delta\phi(t).
\label{grad2}
\end{align}

Let's now take a closer look on $\nabla\Delta\phi(t)$. Note that
\begin{align*}
\phi(t) &= \arctan\left(\frac{\ww\TT \mathbf{x_h}(t)}{\ww\TT \mathbf{x}(t)}\right) \text{\quad or } \nonumber \\
\phi(t) &= \arctan\left(\frac{\ww\TT \mathbf{x_h}(t)}{\ww\TT \mathbf{x}(t)}\right) + \pi.
\end{align*}
%
%
Because of this we can say, if $\ww\TT \mathbf{x}(t) \neq 0$, that $\nabla\phi(t) = \nabla\arctan\left(\frac{\ww\TT \mathbf{x_h}(t)}{\ww\TT \mathbf{x}(t)}\right)$. On the other hand, since $\Delta\phi(t) = \phi(t) - \psi(t)$ and $\psi(t)$ does not depend on $\ww$, we have (we will omit the time dependence for the sake of clarity):
\begin{align*}
\nabla\Delta\phi &= \nabla\phi - \nabla\psi = \nabla\phi = \nabla\arctan\left(\frac{\ww\TT \mathbf{x_h}}   {\ww\TT \mathbf{x}}\right)\\
%
%
&= \frac{\mathbf{x_h}  \cdot \ww\TT \mathbf{x} - \mathbf{x} \cdot \ww\TT \mathbf{x_h} }   {\left[1+\left(\frac{\ww\TT \mathbf{x_h}}{\ww\TT \mathbf{x}}\right)^2\right] \cdot \Big(\ww\TT \mathbf{x}\Big)^2} = \frac{\GG_x(t) \cdot \ww}   {Y^2(t)},
%
%
\end{align*}
where $Y^2(t) = \left(\ww\TT \mathbf{x}(t)\right)^2 + \left(\ww\TT \mathbf{x_h}(t)\right)^2$ is the squared magnitude of the estimated source, and $\GG_x(t) = \mathbf{x_h}(t)  \mathbf{x}\TT(t) - \mathbf{x}(t) \mathbf{x_h}\TT(t)$, thus $\GG_{x_{ij}}(t) = X_i(t)X_j(t) \sin(\phi_i(t)-\phi_j(t))$.

We can now replace $\nabla\Delta\phi(t)$ in \eqref{grad2} to obtain
\begin{align}
\nabla|\varrho|^2 &= \frac{2|\varrho|}{T}\left[\sum_{t=1}^T\frac{\sin[\Phi - \Delta\phi(t)]}{Y^2(t)}\GG_x(t)\right] \ww \nonumber\\
&= 2|\varrho|\left\langle\frac{\sin[\Phi - \Delta\phi(t)]}{Y^2(t)}\GG_x(t)\right\rangle \ww.
\end{align}

\section{Gradient of $J_l$ in IPA}
\label{AppIPAGrad}

In this section we show that the gradient of $J_l$ in Eq. 7 of \cite{Almeida2011} is given by Eq. 8 of \cite{Almeida2011}. Throughout this whole section, we will omit the dependence on the subspace $l$, for the sake of clarity. In other words, we are assuming (with no loss of generality) that only one subspace was found. Whenever we write $\WW$, $\mathbf{y}$, $y_m$ or $\mathbf{z}$, we will be referring to $\WW_l$, $\mathbf{y}_l$, $(\mathbf{y}_l)_{_m}$ or $\mathbf{z}_l$.

The derivative of $\log\left|\det\WW\right|$ is $\WW^{-\mathsf{T}}$. We will therefore focus on the gradient of the first term of Eq. 7 of \cite{Almeida2011}, which we will denote by $P$:
\begin{equation*}
P \equiv \frac{1-\lambda}{N^2} \sum_{m,n} |\varrho_{mn}|^2.
\end{equation*}

Let's rewrite $P$ as $P = \frac{1-\lambda}{N^2}\sum_{m,n} p_{mn}$ with $p_{mn} = |\varrho_{mn}|^2$. Define $\Delta\phi_{mn} = \phi_m - \phi_n$. Omitting the time dependency, we have
\begin{align}
\nabla_{\ww_j}p_{mn}
%
%
&= 2|\varrho_{mn}| \nabla_{\ww_j} \left|\left\langle e^{\textnormal{i}\Delta\phi_{mn}}\right\rangle\right|  \nonumber \\
%
%
%
%
&\hspace{-1cm}= |\varrho_{mn}| \times \Big(\big\langle \cos(\Delta\phi_{mn}) \big\rangle^2 + \textnormal{i}\big\langle \sin(\Delta\phi_{mn}) \big\rangle^2\Big)^{-1/2} \times \nonumber \\
&\times \Big[ 2\big\langle \cos(\Delta\phi_{mn})\big\rangle \nabla_{\ww_j} \big\langle \cos(\Delta\phi_{mn}) \big\rangle + \nonumber \\
&+ 2\big\langle \sin(\Delta\phi_{mn}) \big\rangle \nabla_{\ww_j} \big\langle \sin(\Delta\phi_{mn}) \big\rangle \Big]  \nonumber \\
&\hspace{-1cm}= 2|\varrho_{mn}| \left|\left\langle e^{\textnormal{i}\Delta\phi_{mn}}\right\rangle\right|^{-1} \times \nonumber \\
&\hspace{-1cm} \times \Big[ -\big\langle \cos(\Delta\phi_{mn})\big\rangle \Big\langle\sin(\Delta\phi_{mn}) \nabla_{\ww_j}\Delta\phi_{mn} \Big\rangle + \nonumber \\
&\hspace{-1cm} + \big\langle \sin(\Delta\phi_{mn})\big\rangle \Big\langle\cos(\Delta\phi_{mn}) \nabla_{\ww_j}\Delta\phi_{mn} \Big\rangle\Big],
\label{eq:IPAGrad1}
\end{align}
where we have interchanged the partial derivative and the time average operators, and used 
\begin{equation*}
\Big(\big\langle \cos(\Delta\phi_{mn}) \big\rangle^2 + \textnormal{i}\big\langle \sin(\Delta\phi_{mn}) \big\rangle^2\Big)^{1/2} = \left|\left\langle e^{\textnormal{i}\Delta\phi_{mn}}\right\rangle\right|.
\end{equation*}

Since $\phi_m$ is the phase of the $m$-th measurement, its derivative with respect to any $\ww_j$ is zero unless $m = j$ or $n = j$. In the former case, a reasoning similar to Appendix~\ref{AppRefGrad} shows that
\begin{align}
\nabla_{\ww_j}\Delta\phi_{jk} &\equiv \nabla_{\ww_j}\phi_j - \nabla_{\ww_j}\phi_k = \nabla_{\ww_j}\phi_j = \nonumber\\
&= \frac{[\mathbf{z_h}\cdot\mathbf{z} - \mathbf{z}\cdot\mathbf{z_h}]\cdot\ww_j}{Y_j^2} =  \frac{\GG_z\cdot\ww_j}{Y_j^2},
\end{align}
where $\GG_z(t) = \mathbf{z_h}(t)  \mathbf{z}\TT(t) - \mathbf{z}(t) \mathbf{z_h}\TT(t)$. It is easy to see that $\nabla_{\ww_j}\Delta\phi_{jk} = - \nabla_{\ww_j}\Delta\phi_{kj}$. Furthermore, $p_{mm} = 1$ by definition, hence $\nabla_{\ww_j}p_{mm} = 0$ for all $m \textnormal{ and } j$. From these considerations, the only nonzero terms in the derivative of $P$ are of the form
\begin{align}
\nabla_{\ww_j}p_{jk} &= \nabla_{\ww_j}p_{kj} = 2|\varrho_{jk}|\left|\left\langle e^{\textnormal{i}(\phi_j - \phi_k)}\right\rangle\right|^{-1} \times \nonumber \\
&\hspace{-0.7cm}\times \left[-\big\langle \cos(\Delta\phi_{jk}) \big\rangle \left\langle \sin(\Delta\phi_{jk}) \frac{\GG_z\cdot\ww_j}{Y_j^2} \right\rangle + \nonumber \right.\\
&\hspace{-0.7cm}\left.+\big\langle\sin(\Delta\phi_{jk})\big\rangle \left\langle \cos(\Delta\phi_{jk}) \frac{\GG_z\cdot\ww_j}{Y_j^2} \right\rangle\right].
\label{eq:IPAGrad2}
\end{align}

We now define $\Psi_{jk} \equiv \langle\phi_j-\phi_k\rangle = \left\langle \Delta\phi_{jk} \right\rangle$.
Plugging in this definition into Eq.~\eqref{eq:IPAGrad2} we obtain
\begin{align*}
\nabla_{\ww_j}p_{jk} &= 2|\varrho_{jk}| \times \nonumber \\
& \times \left[-\cos(\Psi_{jk}) \left\langle \sin(\Delta\phi_{jk}) \frac{\GG_z\cdot\ww_j}{Y_j^2} \right\rangle + \nonumber \right.\\
&\left.+\sin(\Psi_{jk}) \left\langle \cos(\Delta\phi_{jk}) \frac{\GG_z\cdot\ww_j}{Y_j^2} \right\rangle\right] \\
&\hspace{-1cm}= 2|\varrho_{jk}| \left[\left\langle -\cos\Psi_{jk} \sin\Delta\phi_{jk} \frac{\GG_z\cdot\ww_j}{Y_j^2} \right\rangle + \nonumber \right.\\
&\left.+\left\langle \sin\Psi_{jk} \cos\Delta\phi_{jk} \frac{\GG_z\cdot\ww_j}{Y_j^2} \right\rangle\right] \\
&\hspace{-1cm}= 2|\varrho_{jk}| \left\langle\sin \left(\Psi_{jk}-\Delta\phi_{jk}\right) \frac{\GG_z}{Y_j^2}\right\rangle\cdot\ww_j,
\end{align*}
where we again used $\sin(a-b) = \sin a \cos b - \cos a \sin b$ in the last step. Finally,
\begin{align*}
\nabla_{\ww_j}P &= \frac{1-\lambda}{N^2}\sum_{m,n} \nabla_{\ww_j}p_{mn} = 2\frac{1-\lambda}{N^2}\sum_{m < n} \nabla_{\ww_j}p_{mn} = \\
= 4\frac{1-\lambda}{N^2} &\sum_{k=1}^N |\varrho_{jk}| \left\langle\sin \left[\Psi_{jk}-\Delta\phi_{jk}(t)\right] \frac{\GG_z(t)}{Y_j(t)^2}\right\rangle\cdot\ww_j.
\end{align*}
which is Eq. 8 of \cite{Almeida2011}.

\section{Gradient of $J$ in pSCA}
\label{AppPSCAGrad}

In this section we derive Eq. 10 of \cite{Almeida2011} for the gradient of $J$. Recall that $J$ is given by
\begin{equation*}
J \equiv \sum_{j=1}^P \left|\sum_{i=1}^N u_{ij}\right| = \sum_{j=1}^P \left|\sum_{i=1}^N \sum_{k=1}^P v_{ik} w_{kj} \right|
\end{equation*}
where the $w_{kj}$ are real coefficients that we want to optimize and the $v_{ik}$ are fixed complex numbers. Also recall that $\textnormal{Re}(.)$ and $\textnormal{Im}(.)$ denote the real and imaginary parts.

We begin by expanding the complex absolute value:
\begin{align*}
\sum_j\left|\sum_{i,k} v_{ik} w_{kj} \right| &= \sum_j \left\{\Bigg[ \textnormal{Re}\Bigg( \sum_{i,k}v_{ik} w_{kj} \Bigg)\Bigg]^2\right.\\
&\left.+ \Bigg[ \textnormal{Im}\Bigg( \sum_{i,k}v_{ik} w_{kj} \Bigg)\Bigg]^2\right\}^{1/2}.
\end{align*}
When computing the derivative in order to $w_{kj}$, only one term in the leftmost sum matters. Thus,
\begin{align}
&\frac{\partial J}{\partial w_{kj}} = \nonumber \\
&= 2\left(\Bigg[ \textnormal{Re}\Bigg( \sum_{i,k}v_{ik} w_{kj} \Bigg)\Bigg]^2 + 
\Bigg[ \textnormal{Im}\Bigg( \sum_{i,k}v_{ik} w_{kj} \Bigg)\Bigg]^2\right)^{-1/2} \hspace{-0.4cm}\times \nonumber \\
& \times \left[ \frac{\partial\left[ \textnormal{Re}\left(\sum_{i,k} v_{ik} w_{kj} \right)\right]^2}{\partial w_{kj}} + \frac{\partial\left[ \textnormal{Im}\left(\sum_{i,k} v_{ik} w_{kj} \right)\right]^2}{\partial w_{kj}} \right] \nonumber \\
&= \frac{1}{\left|\sum_i u_{ij}\right|}\left[ \textnormal{Re}\left(\sum_{i,k} v_{ik} w_{kj} \right) \frac{\partial \ \textnormal{Re}\left(\sum_{i,k} v_{ik} w_{kj} \right)}{\partial w_{kj}} \right. + \nonumber \\
& + \left. \textnormal{Im}\left(\sum_{i,k} v_{ik} w_{kj} \right)\frac{\partial \ \textnormal{Im}\left(\sum_{i,k} v_{ik} w_{kj} \right)}{\partial w_{kj}} \right].
\label{eq:AppPSCAGrad1}
\end{align}

In the sums inside the derivatives, the sum on $k$ can be dropped as only one of those terms will be nonzero. Therefore,
\begin{align*}
\frac{\partial \ \textnormal{Re}\Big(\sum_{i,k} v_{ik} w_{kj} \Big)}{\partial w_{kj}} &= \frac{\partial \ \textnormal{Re}\Big(\sum_i v_{ik} w_{kj} \Big)}{\partial w_{kj}} \\
= \frac{\partial \ \textnormal{Re}\left(\sum_i v_{ik}\right) w_{kj}}{\partial w_{kj}} &= \textnormal{Re}\left(\sum_i v_{ik}\right) = \textnormal{Re} \left(\bar{v}_k\right),
\end{align*}
where we used $\bar{v}_i \equiv \sum_k v_{ki}$ to denote the sum of the $i$-th column of $\VV$. Similarly,
\begin{equation*}
\frac{\partial \ \textnormal{Im}\Big(\sum_{i,k} v_{ik} w_{kj} \Big)}{\partial w_{kj}} = \textnormal{Im} \left(\bar{v}_k\right).
\end{equation*}
These results, with the notation $\bar{u}_j \equiv \sum_k v_{jk}$ as the sum of the $j$-th column of $\UU$, can be plugged into Eq.~\eqref{eq:AppPSCAGrad1} to yield
\begin{equation*}
\mathbf{G}_{kj} = \frac{\partial J}{\partial w_{kj}} = \frac{1}{\left|\bar{u}_j\right|} \Big[ \textnormal{Re}(\bar{v_k})\times\textnormal{Re}(\bar{u}_j) + \textnormal{Im}(\bar{v_k})\times\textnormal{Im}(\bar{u}_j) \Big].
\end{equation*}

\section{Mean field}
\label{AppMeanfield}

In this section we derive Eq. 9 of \cite{Almeida2011} for the interaction of an oscillator with the cluster it is part of. We will assume that there are $N_j$ oscillators in this cluster, coupled all-to-all with the same coupling coefficient $\kappa$, and that all inter-cluster interactions are weak enough to be disregarded. We begin with Kuramoto's model (Eq. 1 of \cite{Almeida2011}) omitting the time dependency:
\begin{align*}
\dot{\phi}_i &= \omega_i + \sum_{k \in c_j} \kappa_{ik}\sin(\phi_k - \phi_i) + \sum_{k \notin c_j} \kappa_{ik}\sin(\phi_k - \phi_i)\\
\dot{\phi}_i &= \omega_i + \sum_{k \in c_j} \kappa_{ik}\sin(\phi_k - \phi_i)\\
&= \omega_i + \sum_{k \in c_j} \kappa_{ik} \frac{e^{\ii(\phi_j-\phi_i)}-e^{-\ii(\phi_j-\phi_i)}}{2\ii}\\
&= \omega_i + \frac{e^{-\ii\phi_i}}{2\ii}\sum_{k \in c_j} \kappa_{ik}e^{\ii\phi_k} - \frac{e^{\ii\phi_i}}{2\ii}\sum_{k \in c_j} \kappa_{ik}e^{-\ii\phi_k}.
\end{align*}
We now plug in the definition of mean field $\varrho_{c_j} e^{\ii\Phi_{c_j}} = \frac{1}{N_j} \sum_{k \in c_j} e^{\ii\phi_k}$ to obtain
\begin{align*}
\dot{\phi}_i &= \omega_i + N_j\frac{e^{-\ii\phi_i}}{2\ii}\kappa\varrho_{c_j} e^{\ii\Phi_{c_j}} - N_j\frac{e^{\ii\phi_i}}{2\ii}\kappa\varrho_{c_j} e^{-\ii\Phi_{c_j}}\\
&= \omega_i + N_j\kappa\varrho_{c_j}\left[ \sin(\Phi_{c_j} - \phi_i) - \sin(\phi_i - \Phi_{c_j}) \right]\\
&= \omega_i + 2N_j\kappa\varrho_{c_j} \sin(\Phi_{c_j} - \phi_i).
\end{align*}

\bibliographystyle{IEEEtran}

\end{document}